# Funnel Transform for Straight Line Detection

Qian-Ru Wei [1,2], Da-Zheng Feng [1,*] *Member IEEE, and* Wei-Xing Zheng[3], *Fellow IEEE*

[1] National Laboratory of Radar Signal Processing, Xidian University, Xi'an, China.

Email: dzfeng@xidian.edu.cn

[2] School of Software and Microelectronics, Northwestern Polytechnical University, Xi'an, China.

Email: weiqianru@nwpu.edu.cn

[3] School of Computing, Engineering and Mathematics, University of Western Sydney, Penrith NSW 2751, Australia.

Email: w.zheng@uws.edu.au

**Abstract:** Most of the classical approaches to straight line detection only deal with a binary edge image and need to use 2D interpolation operation. This paper proposes a new transform method figuratively named as funnel transform which can efficiently and rapidly detect straight lines. The funnel transform consists of three 1D Fourier transforms and one nonlinear variable-metric transform (NVMT). It only needs to exploit 1D interpolation operation for achieving its NVMT, and can directly handle grayscale images by using its high-pass filter property, which significantly improves the performance of the closely-related approaches. Based on the slope-intercept line equation, the funnel transform can more uniformly turn the straight lines formed by ridge-typical and step-typical edges into the local maximum points (peaks). The parameters of each line can be uniquely extracted from its corresponding peak coordinates. Additionally, each peak can be theoretically specified by a 2D delta function, which makes the peaks and lines more easily identified and detected, respectively. Theoretical analysis and experimental results demonstrate that the funnel transform has advantages including smaller computational complexity, lower hardware cost, higher detection probability, greater location precision, better parallelization properties, stronger anti-occlusion and noise robustness.

## I. Introduction

Straight line detection is a classic moderate-level problem in image processing, pattern recognition, and computer vision, and has found extensive applications in extracting highways, buildings, bridges, pipelines, and other line structures from an image. The most famous method for detecting straight lines in digital images is the Standard Hough Transform (SHT) [1]. However, the computational complexity and storage requirement are the two bottlenecks of SHT [2]. Additionally, reference [3] has pointed that the detection accuracy of SHT is sensitive to the resolution of accumulator cell and the detection performance of SHT is easily influenced by noise. To overcome the drawbacks of SHT, many modified versions of SHT have been published [4-7]. In this paper, SHT and its variations are called as Hough Transform (HT)-class methods. Generally, the HT-class methods consist of three basic steps: 1) edge-pixel extraction: edge-pixels are extracted from a grayscale image by edge detectors [8-10] and then form a binary edge image; 2) voting accumulation: according to the certain voting scheme, all or part of the extracted edge-pixels are voted to the accumulator cells in the parameter space; 3) peak selection: the accumulator cells having the local maxima are selected as the line candidates. Furthermore, most of the HT-class methods also include a line verification step trying to minimize the number of false lines. In this paper, the local maximum cell is referred to as a peak or a salient point. Especially, since it is very difficult to accurately find a peak from the flat shape of SHT, a good HT-class method should produce as sharp peaks as possible.

Depending on the voting scheme, the HT-class methods are categorized as *non-probabilistic HT* and *probabilistic HT* [6]. The characteristic of the *non-probabilistic HT* is that all the extracted edge-pixels must participate in the voting accumulation process. The representative SHT [1], fast HT [11], adaptive HT [4,12], and multiresolution HT [13], and so on [14-17] belong to the *non-probabilistic HT*. Besides, since SHT is mathematically equivalent to the Radon transform [18-20], a fast implementation version of Radon transform named as Fourier-based HT (FHT) [19] is also classified as the *non-probabilistic HT*. The FHT skillfully uses the central-slice theorem to finish its voting accumulation process. Hence, FHT can simply extract straight lines by performing a two-dimensional (2D) Fourier transform, a Cartesian-to-Polar coordinates mapping and a one-dimensional (1D) Fourier transform in turn. FHT saves a lot of computational time via the fast Fourier transform operation. However, Shi et al [20] pointed out that the 2D image interpolation involved in the Cartesian-to-Polar coordinates mapping always requires an enormous computational complexity for achieving sufficient detection accuracy, which slows down the speed of FHT to some extent. In addition, such a 2D image interpolation also results in larger artificial errors than 1D image interpolation.

Leandro et al. [21] presented an improved voting scheme of SHT in order to achieve real-time performance. This typical detection scheme, named as Kernel-based HT (KHT), operates on the clusters of approximately collinear edge-pixels. KHT first groups all detected edge-pixels into several clusters. Then the best-fitting line for each cluster is found. The uncertainty of each best-fitting line is computed and determines an elliptical-Gaussian kernel. Finally, KHT votes for all the elliptical-Gaussian kernels. Relative experimental results showed that the use of elliptical-Gaussian kernel significantly improves the performance of voting scheme and produces a much cleaner parameter space. Consequently, KHT makes peaks easy to identify.

The characteristic of the *probabilistic HT* approaches is that they should randomly select a subset of edge-pixels to participate in the voting step [6]. In the *probabilistic HT*, the most representative one is the Randomized HT (RHT) [22] proposed by Xu et al. in 1990. In most case, RHT has low computational complexity. However, it was shown in references [6, 18] that RHT may be difficult to deal with complex and noisy images. Furthermore, it was also indicated in references [6, 18] that RHT sometimes cannot properly detect short lines due to its random sampling strategy. The *probabilistic* approaches can also be applied to the probabilistic Hough Transform (ProbHT) [23] which is different from RHT in both the random sampling strategy and the accumulation method. ProbHT usually uses a small randomly selected subset of edge pixels in accumulation, but this subset has to be quite large to ensure that ProbHT obtains accuracy close to SHT [6]. Random selection of the edge-pixels reduces the computing time of the *probabilistic HT*. However, literature [6] indicated that the performances of the *probabilistic HT* approaches are primarily determined by the number of lines in an image and the noise level. If one of those becomes large, the computing time will increase rapidly.

According to the different way of line parameterizations, the HT-class methods can also be categorized as slope-interpret parameterization [24], $\rho-\theta$ (rho-theta) parameterization [1], circle parameterization [25] and so on [26], in which $\rho-\theta$ parameterization is the most commonly used one. Due to the limitation of the manuscript length, those seldom-used ways will be no longer described in detail.

Except for the HT-class methods, there exist a lot of non-HT ones [27- 31]. Under many cases, the non-HT methods are well adapted for finding short lines rather than long lines. This is because most of the non-HT methods focus their attention on the use of local pixel information, and do not exploit global image information [18]. As we have known, some of the non-HT methods utilize the local statistical knowledge to identify straight lines. For example, Lee et al. [27] described a principal component analysis method for detecting straight lines. Others of the non-HT class methods exploit the local structure information of the edges. For example, Liu and Feng [28] introduced a short line segment detection method. Some of the non-HT class methods start from the pixel information to find out the short lines of an image. For example, Burns et al. proposed a linear-time line detection method with using only gradient orientations [29]. In 2008, Desolneux et al. [30] introduced another non-HT class method which succeeded in controlling the number of false positive detection. Based on an improved Burns method, and combined with a validation criterion inspired from the Desolneux detector, another linear-time line segment detector (LSD) was developed in [31]. By taking a region grouping algorithm, LSD first divides an image into several regions according to the gradient information. Then a rectangular approximation algorithm is used to find a straight line from each region. Every line is judged by the validation criterion in order to exclude non-line structures. LSD has advantages of fast speed, high detection accuracy and very low false positive rate. However, like some non-HT methods,

LSD may not well extract long lines and intersecting lines since it does not use the global image information [18].

In this paper, we establish a new straight line detection method which mainly relies on the proposed funnel transform. The funnel transform method detects straight lines by transforming them into several local maximum points (peaks). The straight lines in the image space are associated with the peaks in the parameter space. The parameter space of the funnel transform is immediately achieved by performing two times 1D Fourier transforms, one time nonlinear variable-metric transform (NVMT) and one time 1D inversion Fourier transform. NVMT is achieved by a simple 1D interpolation operation. Reference [37] suggests that the 1D interpolation operation can be more accurately implemented with two times 1D Fourier transforms. Therefore, the funnel transform can be efficiently and rapidly achieved by only five times 1D Fourier transforms.

The funnel transform based straight line detection (FT-SLD) method has several advantages. 1) Most of the HT or non-HT class methods are only suitable to deal with binary edge maps. Edge extraction process included before line detection increases the computational complexity. What is worse, the detection result of the HT or non-HT methods may be constrained by the performance of edge detection operators. In contrast, our straight line detection method can be directly applied to handle the real grayscale images. Directly performing on the grayscale images makes that there is no need to extract edge pixels from a highly complex image and in a well-designed way, which can reduce the computational cost and avoid the extra error caused by edge operators. 2) The funnel transform method is based on the slope-intercept line equation. Although the linear parameterization increases the difficulty for the funnel transform to create the parameter space, it was shown in reference [32] that the linear parameterization is suitable for hardware implementation and superior to the rho-theta parameterization. On the other hand, the funnel transform can be achieved by five times 1D Fourier transforms. Hence, the hardware cost of funnel transform is relatively low. 3) By using 1D fast Fourier transform, a dramatic reduction in the computation time of the funnel transform is achieved. Further, since the funnel transform separately processes each row or column of an input image, it is easy to be parallelized. Hence, the total elapsed time is significantly reduced. 4) The funnel transform satisfactorily generates the parameter space without the requirement to manually tune parameters for each image. 5) Theoretical analyses show that the peaks of the funnel transform can be theoretically specified by a 2D delta function, while the peaks produced by HT methods are the delta functions only along coordinate rho. In other words, the funnel transform produces more sharp peaks than the HT and non-HT methods, thus making line detection become easier. 6) In particular, experimental results demonstrate that the funnel transform method has good noise immunity, high detection ability, great localization precision and resistance to occlusion.

The remainder of this paper is organized as follows. In Section II, the related work is briefly described. Section III introduces the theory of the funnel transform. A detailed description of the FT-SLD detection method is provided in Section

IV. Section V discusses the computational complexity of several relative straight line detection methods. The experimental results are reported and analyzed in Section VI. Finally, Section VII concludes the paper.

## II. Related work

The SHT [1] is developed for detecting straight lines. Since SHT deals only with binary edge images, we must perform a binary edge extraction process before SHT is applied, which not only increases the computational complexity but also partly decreases the performance of SHT. Let $I_b(x, y)$ be a binary edge image defined in the $x-y$ plane, where $x$ and $y$ are Cartesian coordinates. For the sake of a clear distinction, in this paper, symbol $I$ is always used to denote an image. The subscripts 'b' and 'g' denote the binary and grayscale images, respectively. As shown in equation (1), in a 2D space, the normal equation of a straight line can be parameterized by angle-radius

$$\rho = x\cos\theta + y\sin\theta \tag{1}$$

where $\rho$ and $\theta$ are polar coordinates [1]. Symbol $\rho \in [-R, R]$ is the distance from the origin to the line, where $R = \sqrt{w^2 + h^2}/2$, $w$ and $h$ are the width and height of $I_b(x, y)$, respectively; $\theta \in [0°, 180°)$ represents the line orientation. It is easy to see from equation (1) that a given point $(x_0, y_0)$ in the $x-y$ plane corresponds to a sinusoidal curve $\rho = x_0 \cos\theta + y_0 \sin\theta$ in the $\rho-\theta$ plane. Therefore, the collinear points in the $x-y$ plane along the line $\rho_0 = x\cos\theta_0 + y\sin\theta_0$ will be mapped into several sinusoidal curves in the $\rho-\theta$ plane. Fortunately, these sinusoidal curves intersect at the point $(\rho_0, \theta_0)$.

References [18-20] pointed out that SHT can be deduced from an equivalent transform called Radon transform. The famous Radon transform of $I_b(x, y)$ is usually expressible as

$$\lambda(\rho, \theta) = \mathbf{R}_2\{I_b\} = \iint_D I_b(x, y)\delta(\rho - x\cos\theta - y\sin\theta)dxdy \tag{2}$$

where $D$ is the support region spanned by the variables of integration, $\delta(\bullet)$ indicates a 1D delta function in 2D space, and $\mathbf{R}_2$ denotes the Radon transform operator. The operator $\mathbf{R}_2$ maps $I_b(x, y)$ with axes $x-y$ into Radon space with axes $\rho-\theta$. Let axes $x-y$ be rotated by transformation

$$x = x'\cos\theta - y'\sin\theta \tag{3a}$$

$$y = x'\sin\theta + y'\cos\theta. \tag{3b}$$

And inserting (3) into (2), we have another form of Radon transform as follows

$$\lambda(\rho,\theta) = \mathbf{R}_2\{I_b\} = \iint_D I_b(x'\cos\theta - y'\sin\theta, x'\sin\theta + y'\cos\theta)\delta(\rho - x')dx'dy'$$
$$= \int_{-\infty}^{+\infty} I_b(\rho\cos\theta - y'\sin\theta, \rho\sin\theta + y'\cos\theta)dy' \qquad (4)$$

It is seen from (4) that by summing up pixel values along a line in the image, Radon transform [33] maps a line into a peak. For example, an ideal line image with certain parameters $(\rho_0, \theta_0)$ can be modeled by the 1D delta function

$$I_b(x,y) = \delta(\rho_0 - x\cos\theta_0 - y\sin\theta_0). \qquad (5)$$

Substituting (5) into (4), we have the Radon transform of an ideal line image as follows

$$\lambda(\rho,\theta) = \int_{-\infty}^{+\infty} \delta(\rho_0 - (\rho\cos\theta - y'\sin\theta)\cos\theta_0 - (\rho\sin\theta + y'\cos\theta)\sin\theta_0)dy'$$
$$= \int_{-\infty}^{+\infty} \delta(\rho_0 - \rho\cos(\theta - \theta_0) + y'\sin(\theta - \theta_0))dy'$$
$$= \int_{-\infty}^{\infty} \frac{1}{|\sin(\theta - \theta_0)|} \delta\left(\frac{\rho_0 - \rho\cos(\theta - \theta_0)}{\sin(\theta - \theta_0)} + y'\right)dy'$$
$$= \begin{cases} \frac{1}{|\sin(\theta - \theta_0)|} & \text{for } \sin(\theta - \theta_0) \neq 0 \\ 0 & \text{for } \sin(\theta - \theta_0) = 0 \text{ and } \rho \neq \rho_0 \\ \int_{-\infty}^{\infty} \delta(0)ds & \text{for } \sin(\theta - \theta_0) = 0 \text{ and } \rho \neq \rho_0 \end{cases}$$
$$= \begin{cases} \frac{1}{|\sin(\theta - \theta_0)|} & \text{for } \sin(\theta - \theta_0) \neq 0 \\ \delta(\rho - \rho_0) & \text{for } \sin(\theta - \theta_0) = 0 \end{cases} \qquad (6)$$

Equation (6) demonstrates that Radon transform will map an ideal line image into a butterfly-shaped peak that is not an ideal 2D delta function.

By using the central slice theorem, reference [19] has achieved the fast implementing of Radon transform. As we known, the 2D Fourier transform of $I_b(x,y)$ is given by

$$\tilde{I}(\omega_x, \omega_y) = \mathbf{F}_2\{I_b\} = \iint_D I_b(x,y)\exp(-j\omega_x x)\exp(-j\omega_y y)dxdy \qquad (7)$$

where $\mathbf{F}_2$ denotes the operator of the 2D Fourier transform. Similarly, the 1D Fourier transform of $\lambda(\rho, \theta)$ is given by

$$\mathbf{F}_1\{\lambda\} = \int \lambda(\rho, \theta)d\rho$$
$$= \int [\iint_D I_b(x,y)\delta(\rho - x\cos\theta - y\sin\theta)dxdy]\exp(-j\omega_\rho \rho)d\rho$$
$$= \iint_D I_b(x,y)\left[\int \delta(\rho - x\cos\theta - y\sin\theta)\exp(-j\omega_\rho \rho)d\rho\right]dxdy \qquad (8)$$
$$= \iint_D I_b(x,y)\exp(-j\omega_\rho x\cos\theta - j\omega_\rho y\sin\theta)dxdy$$
$$= \tilde{I}(\omega_\rho \cos\theta, \omega_\rho \sin\theta)$$

where $\mathbf{F}_1$ is the operator of the 1D Fourier transform. Comparing equations (7) with (8) and making transforms $\omega_x = \omega_\rho \cos\theta$ and $\omega_y = \omega_\rho \sin\theta$, then we can easily obtain the following relation

$$\mathbf{F}_2\{I_b\} = \mathbf{F}_1\{\mathbf{R}_2\{I_b\}\}. \tag{9}$$

Equation (9) means that if we first take the 2D Fourier transform of $I_b(x,y)$, second perform a Cartesian-to-Polar coordinates mapping and finally take the 1D Inverse Fourier transform on the re-sampled Fourier field, then Radon transform of $I_b(x,y)$ is achieved quickly. The fast version of Radon transform attains quick implementation because it takes advantage of the computational saving by using the Fast Fourier transform. The Cartesian-to-Polar coordinates mapping requires that each pixel on the re-sampled Fourier field must be obtained by interpolating its location on the $x-y$ grid, and hence the 2D interpolating density is critical in producing accurate detection results. However, 2D complex-valued interpolation [34] not only requires the enormous computational load, but also causes the larger artificial errors. Hence, Ho et al. [19] took an interpolating method, which combines the bilinear interpolation of the four nearest neighbors, for achieving minimal computational cost. The bilinear interpolation is fast, but has lower accuracy. If some higher order interpolation functions [34,35], such as cubic splines [35], are used to replace the bilinear interpolation, then the interpolation accuracy of FHT will be improved at the expense of larger computational load.

## III. Theory of Funnel Transform

### 3.1 Funnel Transform

In this section we propose an interesting transform, named as funnel transform, which consists of four operations on a 2D real grayscale image $I(x,y)$. First, the 1D Fourier transform of $I(x,y)$ along $y$ is marked as $\bar{I}(x,\omega_2)$. Because $I(x,y)$ is real, there is relation $\bar{I}(x,-\omega_2) = \bar{I}^*(x,\omega_2)$, where superscript "$*$" denotes complex conjugate. Relation $\bar{I}(x,-\omega_2) = \bar{I}^*(x,\omega_2)$ can help us reduce the computational complexity of the relative algorithms.

Second, performing a nonlinear variable-metric transform (NVMT) $x' = (\omega_2/\omega_{max})x$ on $\bar{I}(x,\omega_2)$, where $\omega_{max}$ is an angle frequency large enough, then we get a variable-metric image marked as $\bar{I}(x',\omega_2)$. It is worth pointing out that when $\omega_2 < 0$, the formula $x' = (\omega_2/\omega_{max})x = -(|\omega_2|/\omega_{max})x$ not only includes NVMT but also has a reversal operation along $\omega_2$. The reversal operation implies that if $\omega_2 < 0$, then an image pixel corresponding to positive $x$ is mapped into that corresponding to negative $x'$, while an image pixel associated with negative $x$ is mapped into that associated with positive $x'$. In fact, we first perform the NVMT on $\bar{I}(x,\omega_2)$ for $\omega_2>0$ and get $\bar{I}(x',\omega_2)$ for $\omega_2>0$. Then, for $\omega_2<0$, considering $\bar{I}(x,-\omega_2)=\bar{I}^*(x,\omega_2)$ and $x'=-(|\omega_2|/\omega_{max})x$, we have an important relation $\bar{I}(x',\omega_2)=\bar{I}^*(-x',-\omega_2)$, which shows that the NVMT needs only to perform on the top half part of $\bar{I}(x,\omega_2)$. As shown in Fig. 1, if $\bar{I}(x,\omega_2)$ has a rectangular support region, then the core part (NVMT) of the proposed transform is to let this rectangular support region be compressed into a double triangular one possessed by $\bar{I}(x',\omega_2)$. Hence, we will figuratively refer to the proposed transform as funnel one. In addition, it is seen along $\omega_2$ direction that since the NVMT removes the more low frequency components, it is also equivalent to a high-pass filter. It is well-known that a high-pass filter has the ability for enhancing image edges. This is the reason why funnel transform does not require edge extraction process and can directly deal with grayscale images.

Third, the 1D Fourier transform of $\bar{I}(x',\omega_2)$ along $x'$ is performed and $\bar{I}(\omega_1,\omega_2)$ is obtained.

Fourth, the 1D Inverse Fourier transform of $\bar{I}(\omega_1,\omega_2)$ along $\omega_2$ is performed, which gets $\bar{I}(\omega_1,y)$.

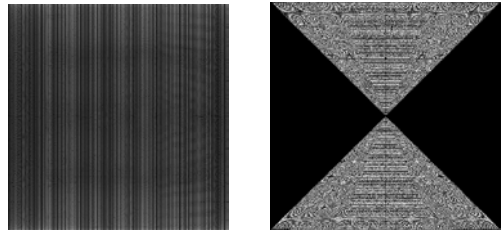

Fig. 1. Rectangular support region is compressed into funnel (double triangular) one by nonlinear variable-metric transform.

Another form of the above funnel transform is named inverse funnel transform as follows. First, the 1D Fourier transform of $I(x,y)$ along $x$ is marked as $\bar{I}(\omega_1,y)$; second, performing the NVMT $y'=(\omega_1/\omega_{max})y$ on $\bar{I}(\omega_1,y)$, then we obtain a variable-metric image $\bar{I}(\omega_1,y')$; Third, the 1D Fourier transform of $\bar{I}(\omega_1,y')$ along $y'$ is performed and marked as $\bar{I}(\omega_1,\omega_2)$. Fourth the 1D Inverse Fourier transform of $\bar{I}(\omega_1,\omega_2)$ along $\omega_1$ is performed and $\bar{I}(\omega_1,\omega_2)$ is converted into $\bar{I}(x,\omega_2)$.

***Remark 1:*** *It is seen that the funnel transform consists of two dual forms in which each is constructed by three 1D Fourier transforms and one NVMT. In the funnel transform, NVMT is generally achieved by a 1D interpolation along direction $x$ or $y$. In particular, 1D interpolation can be highly efficiently implemented by several methods [36-38]. The computational time of these methods [36-38] are nearly equal to each other, which is approximately equal to the total* computational time of two 1D fast Fourier transforms. In contrast, the fast implementation of Radon transform requires one 1D Fourier transform, one 2D Fourier transform and a 2D interpolation. Unfortunately, as we have known, there is not any fast algorithm available for quickly realizing a 2D interpolation.

## 3.2 Point-to-Line Mapping via Funnel Transform

A point-to-line mapping (PTLM) was thoroughly discussed in [39], where it was showed that PTLM must be linear to map collinear points into concurrent lines. Here, we will illustrate that the funnel Transform can automatically implement PTLM. Set the 2D real image $I_g(x,y)$ as a 2D ideal point image $I_g(x,y) = \delta(x-u)\delta(y-v)$ that is equal to infinity and zero at the point $(u,v)$ and at other places, respectively. The 1D Fourier transform of $I_g(x,y)$ along $y$ can be represented as

$$\overline{I}(x,\omega_2) = \delta(x-u)\exp(-j\omega_2 v). \tag{10}$$

It is easy to see that $\overline{I}(x,\omega_2)$ indicates a line image in $x-\omega_2$ coordinates with intercept $u$ and along $\omega_2$. The NVMT of $\overline{I}(x,\omega_2)$ is easily obtained as

$$\overline{I}(x',\omega_2) = \delta(x'-(\omega_2/\omega_{max})u)\exp(-j\omega_2 v). \tag{11}$$

where $\omega_{max}$ denotes a constant and is an angle frequency large enough. The variable-metric image $\overline{I}(x',\omega_2)$ is resampled from a $x-\omega_2$ grid to a $x'-\omega_2$ grid by the NVMT. Very interestingly, the first factor on the right side of equation (11) represents an ideal line image passing zero point and having slope rate $u/\omega_{max}$, while the second factor denotes linear phase $\omega_2 v$. Taking the 1D Fourier transform of $\overline{I}(x',\omega_2)$ along $x'$, we have

$$\overline{I}(\omega_1,\omega_2) = \exp(-(\omega_2/\omega_{max})u\omega_1)\exp(-jv\omega_2) = \exp(-j(v+(u/\omega_{max})\omega_1)\omega_2). \tag{12}$$

Furthermore, carrying out the 1D Inverse Fourier transform of $\overline{I}(\omega_1,\omega_2)$ along $\omega_2$ yields

$$\overline{I}(\omega_1,y) = \delta(y-(v+(u/\omega_{max})\omega_1)). \tag{13}$$

This shows that a point $(u,v)$ in $x-y$ plan is mapped into a line $\overline{I}(\omega_1,y)$ in $\omega_1-y$ coordinates with intercept $v$ and slope rate $u/\omega_{max}$. Appendix A shows that in 3D space, the funnel transform can automatically achieve a point-to-plane mapping.

## 3.3 Line-to-Point Mapping via Funnel Transform

A line-to-point mapping (LTPM) was carefully studied in some recent references [40, 41]. Interestingly, the funnel transform can automatically achieve LTPM. Let us now describe the basic idea of the funnel transform for mapping a straight line into a peak. In $x - y$ plane, the slope-intercept equation of a line is expressible in the linear form as

$$y = kx + b = \tan(\theta)x + b \tag{14}$$

where $k$ and $b$ represent slope rate and intercept distance, respectively, and $\theta$ is the angle between the x-axis and the line. Without losing generality, we assume that the x-axis is along the row direction of an image, since our discussion holds also if the x-axis is along the column direction.

On the basis of the slope-intercept line equation shown in equation (14), we can set the 2D real image $I_g(x, y)$ as a 2D ideal line image, which can be described by

$$I_g(x, y) = \delta(y - k_r x - b_r) \tag{15}$$

where $k_r$ and $b_r$ denote slope rate and intercept distance of the line, respectively. Equation (15) means that the ideal image $I_g(x, y)$ is equal to infinity and zero on the line $y = kx + b$ and at other places, respectively. Performing the 1D Fourier transform of $I_g(x, y)$ along the independent variable $y$ yields within a scale the following relation

$$\begin{aligned} \overline{I}(x, \omega_2) &= \int_{-\infty}^{+\infty} I_g(x, y) e^{-j\omega_2 y} dy = \exp(-j(k_r x + b_r)\omega_2) \\ &= \exp(-jb_r \omega_2) \exp(-jk_r \omega_2 x) \end{aligned} \tag{16}$$

In LTPM, we need to integrate all the frequency-domain signals associated with the single line. However, equation (16) shows that $\omega_2$ and $x$ are coupled each other, which causes a difficulty for integrating the signals of a line. In order to remove this coupled effect, we take the NVMT along $x$ with respect to $\omega_2$ as follows

$$\omega_2 x = \omega_{\max} x' . \tag{17}$$

Substituting equation (17) into equation (16) yields

$$\overline{I}(x', \omega_2) = \exp(-jb_r \omega_2) \exp(-jk_r \omega_{\max} x') . \tag{18}$$

Very interestingly, equation (18) has removed the couple between $\omega_2$ and $x$.

Then, performing 1D Fourier transform and 1D inverse Fourier transform of $\overline{I}(x', \omega_2)$ with respect to the independent variables $x'$ and $\omega_2$, respectively, we have up to complex scale the relation

$$\overline{I}(\omega_1, y) = \delta(\omega_1 + \omega_{max} k_r)\delta(y + b_r). \tag{19}$$

It is observed from equation (19) that via the funnel transform, the ideal line image has been integrated into an ideal sharp peak in the slope-intercept parameter space. Specially, in order to remove the effect of complex number on the parameter space, we usually take the absolute value of $\overline{I}(\omega_1, y)$. More generally, the funnel transform can automatically achieve a plane-to-point mapping in 3D space, which is illustrated in Appendix A.

For the sake of clarity, in Fig. 2 we exhibit the procedures in which the funnel transform maps the lines into sharp peaks.

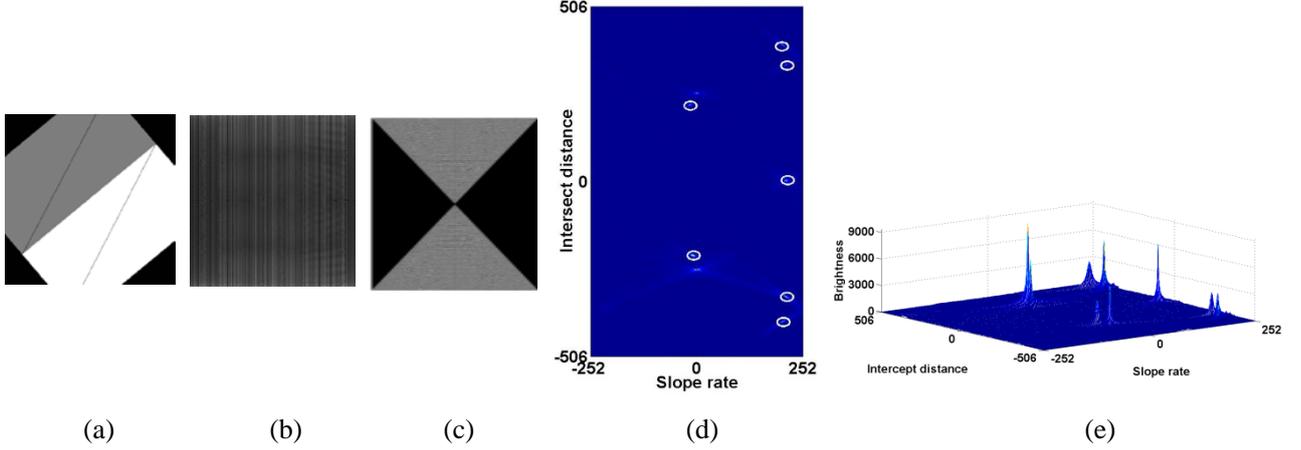

(a)　　　　(b)　　　　(c)　　　　(d)　　　　(e)

Fig. 2. The funnel transform is used to find lines formed by ridge-typical and step-typical edges. The first to the fifth sub-figures respectively denote (a) the original image, (b) 1D Fourier Spectrum, (c) result of NVMT, (d) salient (peaks) points in the parameter space, and (e) 3D visualization of sharp peaks in the parameter space.

***Remark 2:*** *It is found via experiments that a straight line formed by step-typical edge can also be mapped into a sharp peak by the funnel transform. This is in fact supported by the high-pass filtering ability of the NVMT.*

***Remark 3:*** *It is shown from (19) that the funnel transform makes an ideal line image be converted into an ideal point image which is described by 2D delta function. In contrast, the peak of Radon transform cannot be exactly described as any 2D delta function (see equation (6)). Under the most cases, a peak of Radon transform is diffuse and sharp only along a special direction. Therefore, it is difficult to detect a peak of Radon transform.*

Fig. 3 compares the parameter spaces of SHT, FHT and the funnel transform. The image includes three ideal ridge-typical lines with single-pixel width. Funnel transform, SHT and FHT separately perform on Fig. 3(a). In each parameter space, only the peak corresponding to line $L_1$ is plotted. The experiment result shows that the peaks of SHT and FHT are butterfly-shaped, only a peak of the funnel transform is gathered as an obvious salient point. The 3D view of the parameter space indicates that a peak of the funnel transform is sharper than that of SHT and FHT.

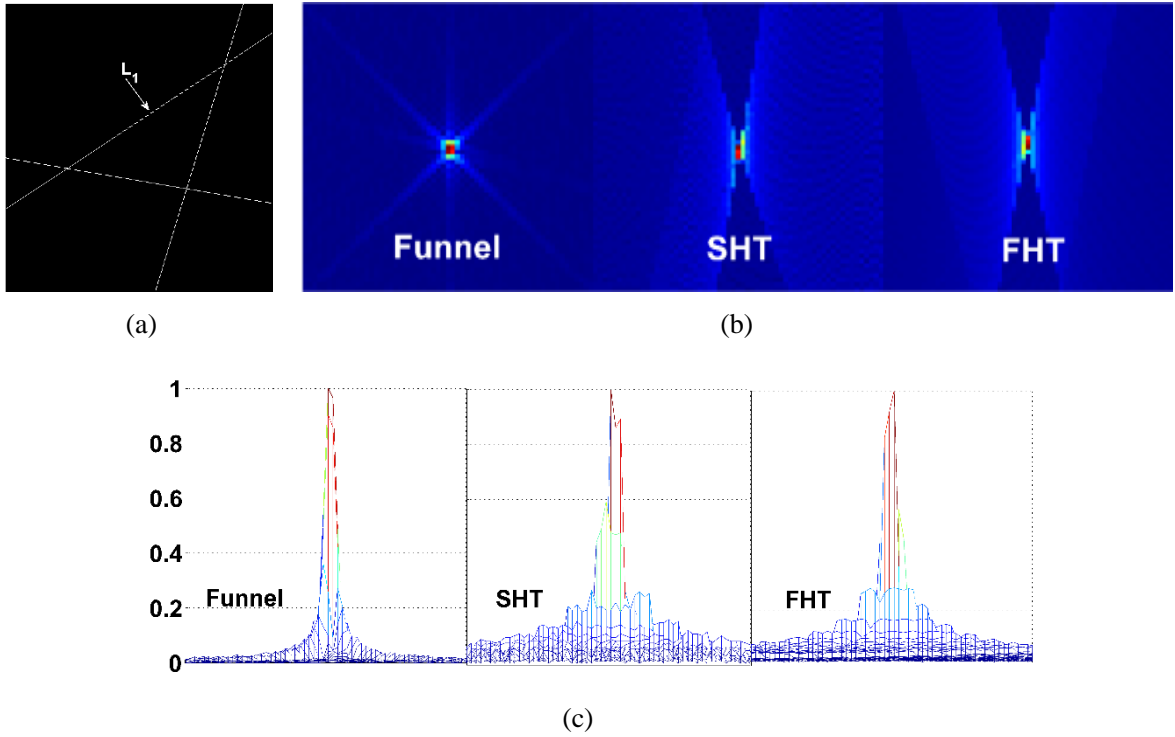

(a)　　　　　　　　　　　　　　(b)

(c)

Fig. 3 Parameter spaces are obtained by the funnel transform and SHT and FHT. (a) Three ideal ridge-typical lines with single width. (b) The peaks corresponding to line $L_1$ obtained by the funnel transform, SHT and FHT. (c) 1D profiles of the peaks.

## IV. Funnel Transform for Straight Line Detection

In digital image processing, the discrete Fourier transform is usually used in the funnel transform. Since the discrete Fourier transform is well-known, its implementation is not discussed here. Moreover, the NVMT is very accurately implemented by several methods such as the fast chirp-z transform [36] and other methods [37, 38]. In this paper, we adopt the fast chirp-z based 1D interpolation with higher interpolation accuracy.

### 4.1 Funnel Transform Frame for Straight Line Detection

The aim of this paper is to propose a *revolutionary* new straight line detection method which adopts the elegant transformation algorithm, the funnel transform proposed in the preceding section. The well-known slope-intercept parameterized lines are transformed by the funnel transform into several peaks in the parameter space. The results obtained by the funnel transform are represented by the coordinates of peaks in the slope and intercept parameter space. Therefore, the location of a straight line on the image is easily determined by the correspondence between the parameters of the straight lines in the image space and the coordinates of the peaks in the parameter space. The proposed straight line detection method consists of three steps (preprocessing, funnel transform and post-processing), which are illustrated by the block diagram shown in Fig. 4.

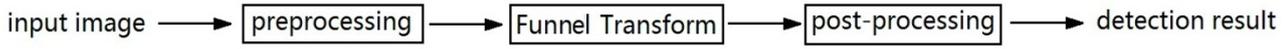

Fig. 4. Block diagram of funnel-transform-based straight line detection method.

Our detection method starts from the preprocessing step. The preprocessing step does not include edge-pixels extraction and noise reduction, but requires the expansion of the input image. The purpose of the preprocessing is to simplify the procedure for distinguishing the true peaks from the false ones. We will explain how the preprocessing is accomplished in the next subsection.

As shown in the above block diagram, the funnel transform is performed on the preprocessed grayscale image, yielding the parameter space. It is well-known that in digital image processing, the very large slope rate and quite great intercept distance of a line cannot be represented as their true values, which results in that they may be periodically wrapped in the parameter space. This wrapping problem leads to the problems of slope ambiguity and intercept ambiguity. It is essentially caused by the periodicity of the 1D Fourier transform. Hence, in order to reduce the ambiguity on slope and intercept and decrease the influence of the wrapping problem on extraction result, according to the slope rate, we divide the parameter space into two dual parts: regular parameter space and inverse parameter space, where the regular parameter space has the slope rate range $(-1,1]$ and the inverse parameter space is within the slope rate range $(-\infty,-1] \cup (1,+\infty)$. The dual parameter spaces are respectively achieved by the funnel transform and the inverse one.

After obtaining two parameter spaces, a post-processing step is used to select true peaks and remove false peaks. It is worth mentioning that some false peaks are actually caused by the periodicity of Fourier transform rather than the noise. Hence, the post-processing step includes peak detection and line verification. The above description can be viewed as an outline of our proposed method. In the following, the detailed descriptions about the first to the third blocks of Fig. 4 will be given.

## 4.2 Preprocessing

It is well-known that lines with very large slope rate and quite great intercept distance can be periodically wrapped by a digital processor. This implies that we cannot use the digital funnel transform to handle all the lines in an image. To overcome the wrapping problem, we expect that each line can be mapped into at least one corresponding peak which has neither slope wrapping nor intercept wrapping. For this purpose, we divide all the lines into two groups according to different slope rates, as indicated in Fig. 5. The lines with slope rate range $(-1,1]$ (slope angle range $\theta \in (-45^o, 45^o]$) are classified as the first group and formulated by slope and *y*-intercept; the lines with slope rate range $(-\infty,-1] \cup (1,+\infty)$ (slope angle range $\theta \in (45^\circ, 135^\circ]$) are classified as the second group and described by inverse-slope and *x*-intercept. The

13    / 35

funnel transform is used to detect the lines in Group One and its dual form, the inverse funnel transform, is exploited to identify the lines in Group Two. Hence, the parameter space for the proposed straight line detection method consists of two parts which are simply referred to as regular space and inversion space. The two dual parameter spaces double the computational load of the detection method. Fortunately, the regular and inversion spaces can be achieved by a parallel processor, which saves the run-time of the proposed detection method.

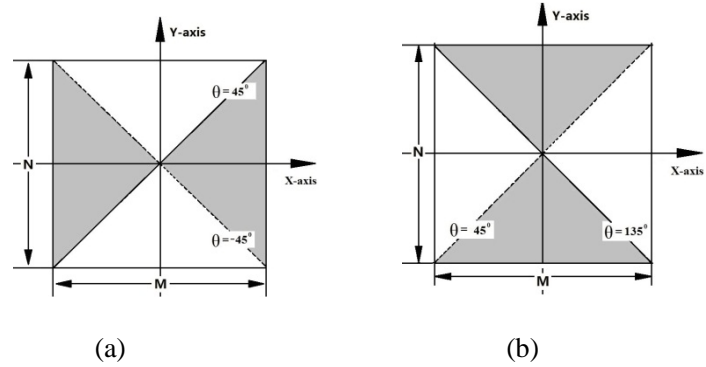

(a)                    (b)

Fig. 5. All lines are divided into two groups. (a) Lines in the first group with slope rate range $(-1,1]$ (slope angle range $\theta \in (-45^o, 45^o]$). (b) Lines in the second group with slope rate range $(-\infty, -1] \cup (1, +\infty)$ (slope angle range $\theta \in (45°, 135°]$).

Without loss of generality, the regular parameter space is mainly considered. We now explain why an input image should be expanded and how the expanded operation is implemented. It is obvious that our discussion holds for both regular and inversion spaces.

For a line with slope range $-1 < k \leq 1$, its corresponding salient (peak) point may suffer from the problem of intercept ambiguity that is caused by periodically wrapping. Appropriately padding zeros along y-axis can allow the lines in the first group to be free from the intercept ambiguity. Assume that an input digital image $I(x,y)$ has size $M \times N$, and the center of Cartesian coordinates is placed at the image center. It is easily known from Fig. 5 that the line having slope $-1$ and passing the pixel at the bottom left-hand corner of the image should achieve the smallest y-intercept $-(N+M)/2$. Similarly, the line having slope 1 and passing the pixel at the top left-hand corner of the image has the biggest y-intercept $(N+M)/2$. Consequently, we should pad $M/2$ row zeros above the first row and behind the last row, for regular parameter space to fully express the lines in the first group. After padding zeros, $I(x,y)$ is extended to the size of $M \times (M+N)$. Performing the funnel transform on the expanded image makes that a line in the first group be mapped into a clearly and easily identifiable salient (peak) point without ambiguities of intercept and slope. Likewise, the problem of intercept and slope ambiguity of lines in the second group can be solved by performing the inverse funnel transform on the expanded image. Further, the another benefit for expanding the image is that the positions of the lines in the original image are more easily found and confirmed by the line verification step.

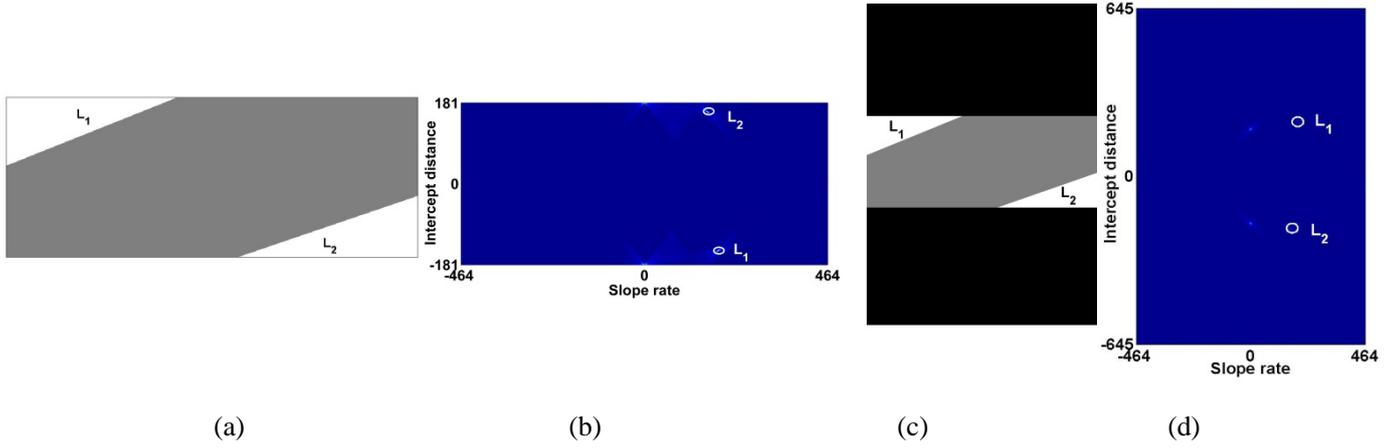

|(a)|(b)|(c)|(d)|

Fig. 6. Perform funnel transform on image with and without expanding. (a) Two lines in input image. (b) The regular parameter space obtained by the funnel transform on the image without expanding. (c) The expanded image. (d) The regular parameter space obtained by the funnel transform on the expanded image.

Figs. 6 and 7 describe the problems of intercept ambiguity and slope ambiguity, respectively. First, we perform the funnel transform separately on an image with and without expanding. As shown in Fig. 6(a), two lines $L_1$ and $L_2$, which are classified as the first group, have positive and negative intercept distances, respectively. Fig. 6(b) is the result of the original image being processed by the funnel transform. Clearly, the salient (peaks) points corresponding to line $L_1$ and line $L_2$ suffer from the problem of intercept wrapping. Generally speaking, if the intercept distance satisfies condition $b_y > N/2$ or $b_y \leq -N/2$, then it is periodically expressed as

$$b_y = Np + \tilde{b}_y \qquad (20)$$

where the wrapping multiplicity $p$ is an integer and the intercept distance $\tilde{b}_y$ in the regular parameter space lies in $(-N/2, N/2]$. If the intercept ambiguity is eliminated by image expanding, then the salient (peaks) points of $L_1$ and $L_2$ are not periodically wrapped as shown in Fig. 6(d). Fig. 6 tells us that expanding image is necessary, even though all the lines lie in the first group. Without expanding image, we must take an extremely complex post-processing algorithm to determine the wrapping multiplicity $p$ of each wrapped salient (peak) point.

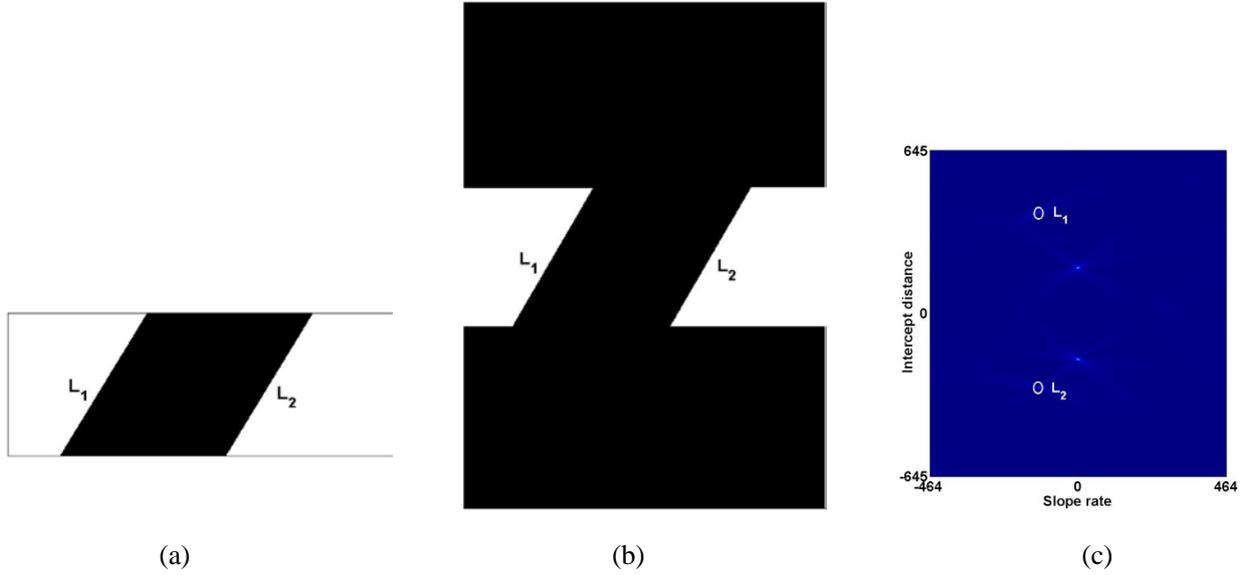

(a) (b) (c)

Fig. 7. (a) Two lines with the same slope rates. (b) expanded image of (a). (c) In the regular parameter space, the salient points corresponding to line $L_1$ and line $L_2$ have been migrated due to the ambiguity of the slope rate.

Fig. 7(a) draws two lines with slope rate $\sqrt{3}$. The funnel transform is performed on the expanded image as shown in Fig. 7(b). The obtained regular parameter space is shown in Fig. 7(c). The salient points formed by the lines are marked with the white circles. However, we find that the salient points marked by the white circles in Fig. 7(c) should correspond to the lines having slope rates $\sqrt{3}-2$. That is to say, the salient (peaks) points of line $L_1$ and line $L_2$ are periodically wrapped into lines with slope rates $\sqrt{3}-2$. Actually, in regular space, all the salient points with the slope wrapping are from the lines in the second group. Generally speaking, if the slope rate $k$ of a line lies within $(-\infty,-1]\cup(1,\infty)$, then it is periodically expressed as

$$k = 2q + \tilde{k} \tag{21}$$

where the wrapping multiplicity $q$ is an integer and the slope rate $\tilde{k}$ in the regular parameter space lies in $(-1,1]$.

By inserting zeros to the image, we can partly eliminate the problem of slope ambiguity. However, no matter how much zeros are inserted into the input image, we cannot completely solve the problem of slope ambiguity since the slope rate situates in $(-\infty,\infty)$. Even worse, over-inserting zeros will lead to that the computation time of the funnel transform is increased considerably. Hence, the way that all the lines have been divided into the two groups according to the slope rate range are used to avoid the problem of slope ambiguity.

It is easily seen that the number of salient (peaks) points in the regular space is more than that of the lines in the input image. There are four salient (peaks) points in Fig. 6(d), but there only are two lines in Fig. 6(a). In Fig. 7, the similar things are generated. This is caused by the adopted zero padding process. For a slope rate range $(-1,1]$, a default padding method means that $M/2$ rows of zeros are padded separately above the first row and behind the last row of the input image. If not specifically emphasizing, we adopt this default padding method, as shown in Fig. 6(c). This default padding method causes the top and bottom artificial boundaries of the original image, and increases two level straight lines. Via the funnel transform, the artificial formed lines are focused into two salient points (peaks).

## 4.3 Relationship between Salient Point Coordinates and Line Parameters

According to the coordinates of a salient point (peak) in the parameter space, it is easy to recover the parameterization information of a straight line in the image space. Fig. 8(a) indicates a straight line $L_1$ with slope angle $\theta$ and y-intercept $b_y$. Fig. 8(c) shows a straight line $L_1$ with inverse slope angle $\theta$ and x-intercept $b_x$. The straight line $L_1$, angle $\theta$ and intercept are marked in Fig. 8 and denoted by red, green and yellow color, respectively. The salient point (peak) in the regular parameter space corresponds to $L_1$ having been marked in Fig. 8(b). Assume that the salient point (peak) mapped by $L_1$ has coordinates $(m,n)$ and the original image has size $M \times N$. If the peak corresponding to line $L_1$ has no ambiguity on the intercept and slope in the regular parameter space, the slope rate and intercept distance of line $L_1$ are respectively computed by

$$k = 2m/M, \text{ for } m = -M/2, -M/2+1, \cdots, M/2. \tag{22}$$

$$b_y = n, \text{ for } n = -(M+N)/2, -(M+N)/2+1, \cdots, (M+N)/2. \tag{23}$$

Similarly, if there does not exist ambiguity on the intercept and slope in inverse parameter space, then a salient (peak) point $(m,n)$ in inverse parameter space determines a straight line with inverse-slope and x-intercept computed respectively by

$$1/k = 2n/N, \text{ for } n = -N/2, -N/2+1, \cdots, N/2 \tag{24}$$

$$b_x = m, \text{ for } m = -(M+N)/2, -(M+N)/2+1, \cdots, (M+N)/2. \tag{25}$$

It is worth to note that the straight line $L_1$ shown in Fig. 8 has slope rate smaller than 1. Hence, from the inverse parameter space shown in Fig. 8(d), we do not easily see the salient point (peak) corresponding to $L_1$, because the brightness of the salient (peaks) points is significantly weakened by the problem of slope ambiguity. The white circles in Fig. 8(d) actually correspond to the right and left boundaries of the original image.

Equations (22) and (23) show that for the funnel transform, the intercept resolution is one pixel. The slope resolution is $2/M$, and the inverse-slope resolution is $2/N$. That is to say, the resolution precision of the funnel transform depends on the size of the input image. The larger the input image, the higher the resolution precision of the funnel transform. Of course, inserting zeros can slightly improve the resolution precision of slope rate. However, such precision improvement is generally at the cost of increasing the computational complexity. Furthermore, restricted by the resolution of input image, inserting zeroes cannot freely increase the resolution precision of slope rate. Our extensive experiments show that inserting zeroes can at most enhance about 20% resolution for slope rate and intercept distance.

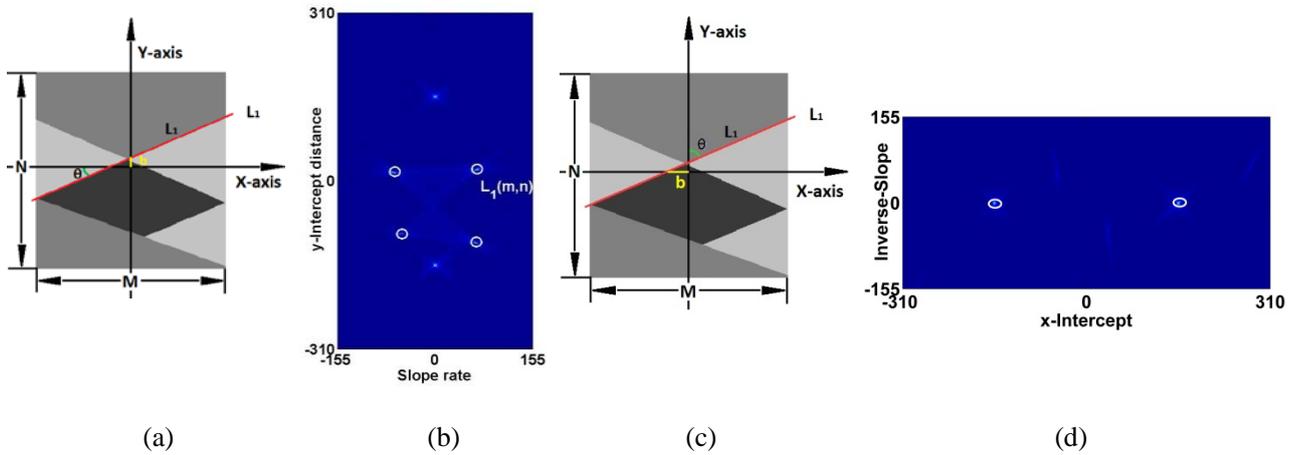

(a)　　　　　　　　(b)　　　　　　　　(c)　　　　　　　　(d)

Fig. 8. (a) Lines represented by slope-intercept form. (b) Peaks shown in the regular parameter space. (c) Lines represented by inverse-slope-intercept form. (d) Peaks shown in the inverse parameter space.

## 4.4 Post-processing: Peak Detection

After performing the funnel transform, lines in the image space are transformed into several salient points (peaks) in the two dual parameter spaces. Here, a list-based peak detection method inspired by Leandro et al. [21] is adopted.

For every point in the regular or inverse parameter space, its value is regarded as its brightness. We establish a candidate list to record all points in the parameter space. The points in the list are sorted in descending order by their brightness. Then the points are visited in recorded order. For the present point, we check if its neighborhood includes any point that has been visited. If so, the checked point should not have the local maximum brightness. In this case, we mark the present point as visited and move to the next point. Otherwise, we add the present point into the list of salient (peaks) points (simply called the detected list), mark it as visited, and then move to the next point in the candidate list. This process continues until the number of points in the detected list reaches a specified number or the visited point has smaller brightness than a predefined threshold $T$. It is worth mentioning that the neighborhood of the visited point is usually taken as size $3 \times 3$. However, if the image is subject to strong noise or has intricate texture, the neighborhood size can be increased appropriately.

Of course, except for the above-mentioned list peak detection method, several other effective peak detection methods [1, 21, 42] have been established. They can replace the list peak detection method. Due to the limited space, these peak detection methods are not described here.

## 4.5 Post-processing: Line Verification

Since the lines in an image are divided into two groups, they must be verified in both regular space and inverse space.

Certainly, not every point in the parameter spaces needs to be verified. Only the points in the detected list should be verified. Those points that need to be verified consist of true peaks, wrapping peaks and false peaks. Here, false peak denotes the salient (peak) point formed by a false line and may be aroused by noise, intricate texture and some other factors. In addition, the peaks associated with the artificial boundaries and caused by expanding the input image are also considered as false peaks.

The wrapping peaks with the ambiguity of slope rate and intercept distance are aroused by the periodical phenomenon of digital processing. As shown in Fig. 9(a), the line $L_1$ with slope rate 3.27 ($\theta = 73°$) is included in an image with size $170 \times 428$. The salient (peak) point marked by a white circle in Fig. 9(b) is a wrapping peak corresponding to $L_1$. Due to the wrapping, the brightness of this peak becomes very dark. In the regular parameter space, the coordinates of this wrapping peak are (−61, −58). According to equations (22) and (23), we compute that this wrapping peak should correspond to a line with slope −0.73 and intercept −58. Clearly, such a line does not exist in Fig. 9(a). It is interesting that the salient (peak) point of line $L_1$ in the inverse parameter space has coordinates (18, 65). According to equations (24) and (25), we can correctly recover the parameterization information of line $L_1$ from the coordinates of this salient (peak) point. This demonstrates that a wrapping peak must be associated with a real line in the image space. And, if a line corresponds to a wrapping peak in regular parameter space, then in inverse parameter space the peak corresponding to this line must not be a wrapping one.

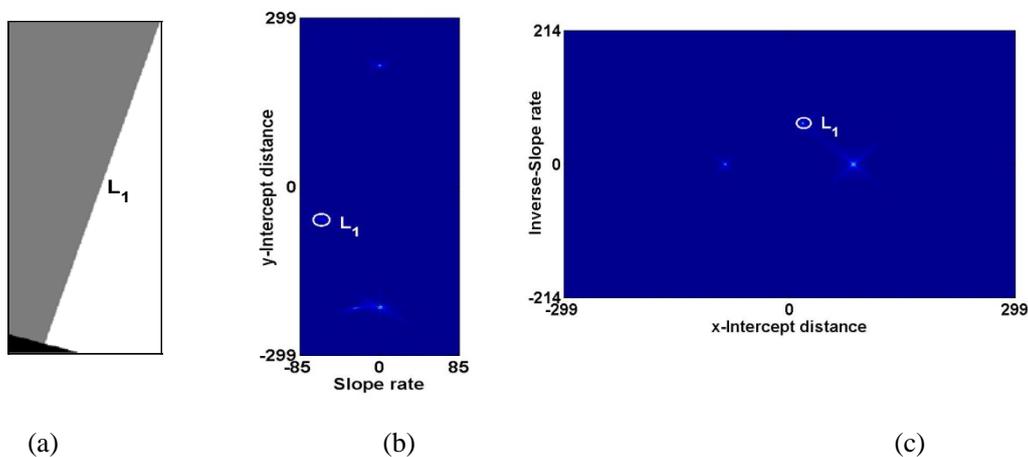

(a) (b) (c)

Fig. 9. Due to ambiguity, salient (peak) points correspond to non-existent lines. (a) Line $L_1$ with slope 3.27. (b) One wrapping peak in the regular parameter space. (c) One unwrapping peak in the inverse parameter space.

A primary function of line verification is to identify all the true salient (peak) points from the detected list. Hence, if a certified point is false, we remove it from the detected list. The wrapping peaks are an inevitable wrapping problem caused by the periodicity of Fourier transform. Evidently, in regular parameter space, wrapping peaks are formed by the lines in the second group; while in inverse parameter space, wrapping peaks are constructed by the lines in the first group. Experiments show that the brightness of a peak not only depends on the length of its corresponding line, but is also influenced by the location and slope rate of the line, the complicated degree and the contrast of the grayscale image. Furthermore, experiments also reveal that the brightness of salient (peaks) points is weakened by the problem of slope ambiguity. Hence, for most images, only a few peaks with the wrapping intercept distances have sufficient brightness such that they exceed the predefined threshold $T$.

Based on the previous analysis and definition, the process of line verification is described as follows.

1) First, for each salient (peak) point in the detected list, the slope and intercept of its candidate line are first computed by its coordinates.
2) Second, in the original image, we construct a long image band with 3~7 pixels width, where the center line of this image band is consistent with the candidate line.
3) Third, linearly integrating this image band along the length direction yields one-dimensional image with 3~7 pixels.
4) Fourth, it is easily determined from this one-dimensional image whether this candidate line is an existing line or not; if there is not this candidate line, then the peak is removed from the detected list.
5) Fifth, if required, we can also compute the width and length of this candidate line from the image band.
6) Repeat steps 1) to 5) until all peaks in the detected list have been verified.

Except for the above line verification method, we can also use several other methods [1, 41]. For example, according to the Helmholtz principle [41], the judgment of a true salient (peak) point can be treated as a simplified hypothesis testing problem, but this is not discussed in this paper.

## V. Computational Complexity Analysis

In most of line detection methods, since the post-processing usually involves relatively small computational volume, their computational complexities usually are ignored. The funnel transform consist of three 1D Fourier transforms and one NVMT. If the input image has size $M \times M$, then the size of two expanded images can be approximated as $2M \times M$ and $M \times 2M$. Along the length direction of the two expanded images, $4M$ times performing 1D Fast Fourier transform with $2M$ points takes the computational complexity $O(4M \times 2M \log 2M)$ that is approximated as $O(8M^2 \log M)$ if $M$ is large enough; along the width direction of the two expanded images, $4M$ times performing 1D Fast Fourier transform with $M$ points takes the computational complexity $O(4M^2 \log M)$. It is easily shown from [36] that two NVMT cost $4M$ times 1D Fast Fourier transforms with $M$ points, and thus take the computational complexity $O(4M^2 \log M)$. Noticeably, if considering the symmetrical relation $\bar{I}(x',\omega_2) = \bar{I}^*(-x',-\omega_2)$, it is only required to perform the NVMT on the top half parts of the two expanded image. Therefore, the computational complexity of the FT-SLD method is about $O(16M^2 \log M)$. If a linear interpolation method is used in the NVMT, then the computational complexity of the NVMT is $O(M^2)$. In this case, the computational complexity of the funnel transform is about $O(12M^2 \log M)$.

The binary edge map of an input image is necessary to most of straight line detection methods such as SHT. Hence, when we compute the computational complexities of these methods, the computation time for obtaining the edge map should be counted. Here, let us take SHT as a typical example. Suppose that SHT utilizes the Canny edge filter [9] to produce a binary edge map. Since the Canny edge filter can be realized by the convolution of the input image with two level- and vertical-derivative kernels, it can be fast implemented by three 2D Fast Fourier transforms and two 2D Inverse Fast Fourier transforms. Hence, to obtain the binary edge image, the Canny operator takes at least $O(10M^2 \log M)$ the computational complexity. Suppose that the achieved edge map has $L$ edge-pixels. When the parameter space of SHT is divided into $M \times M$ accumulator cells, SHT needs $O(2L \times M)$ operations to finish the voting accumulation. The computational complexity of SHT is at least equal to $O(10M^2 \log M) + O(2L \times M)$ which is dependent on the complexity of the image. In most cases, the number of edge-pixels satisfies condition $M << L < M^2$. Consequently, $O(2L \times M)$ may be much bigger than $O(2M^2)$ and even much larger $O(2M^2 \log M)$. Hence, the computational complexity of SHT may be much higher than that of the funnel transform.

The FHT involves one 2D Fourier transform, one 1D Fourier transform and one Cartesian-to-Polar coordinates mapping. The computational complexity of 2D Fourier transform and 1D Fourier transform are $O(2M^2 \log M)$ and $O(M^2 \log M)$, respectively. The Cartesian-to-Polar coordinates mapping involves a 2D image interpolation. The time complexity of 2D interpolation operation depends on the interpolating density that is related to the detection accuracy of FHT. If the bilinear interpolation method is used, then the time complexity of the Cartesian-to-Polar coordinates mapping will be about $O(10M^2)$, where computing interpolation positions takes $O(2M^2)$, and calculating $4M^2$ interpolation coefficients and $M^2$ interpolation values cost $O(8M^2)$. Therefore, the computational complexity of FHT is $O(13M^2 \log M) + O(10M^2)$, where we also include the computational complexity of Canny operator. If a linear interpolation method is used in the NVMT, then the computational complexity of FHT is above that of the funnel transform. If some more accurate 2D interpolation methods [33] are adopted, then the computational complexity of SHT and FHT will be significantly higher than that of the funnel transform.

# VI. Experiment Results

In this section, we analyze and assess the performance of the FT-SLD method in terms of noise resisting capability, anti-occlusion ability, true positive detection rate and localization accuracy. In this paper, all the reported experiments are performed on an Intel Xeon CPU E3-1230 (3.30 GHz) with 8GB RAM. All the artificial images used in the following experiments have been normalized, i.e., their intensities are ranged from 0 to 1. Furthermore, in this paper all experiments will be performed with the same parameter settings regardless of the different image origin, scene, and resolution.

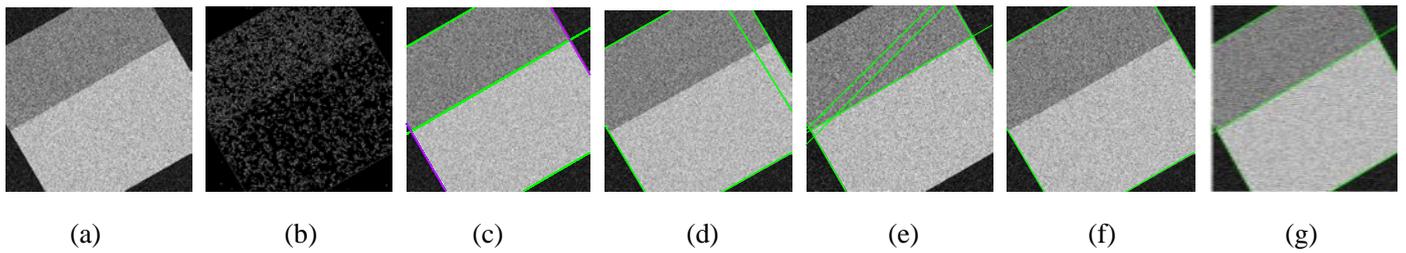

(a)　　　(b)　　　(c)　　　(d)　　　(e)　　　(f)　　　(g)

Fig. 10. Lines are detected from step image with additive Gaussian noise. (a) Image suffers from zero mean Gaussian noise with variance 0.1. (b) Binary edge image obtained by Canny detector. Detection result of (c) funnel transform, (d) KHT, (e) SHT, (f) LSD and (g) FHT. In Fig. 10(b), three parameters required by Canny detector are set as $[T_{low}, T_{high}, \sigma] = [0.14, 0.35, 1]$.

A big sigma value will better smooth the Gaussian noise. Hence, a small sigma is taken. A small sigma provides a clear view of the effect of noise on straight line detection results.

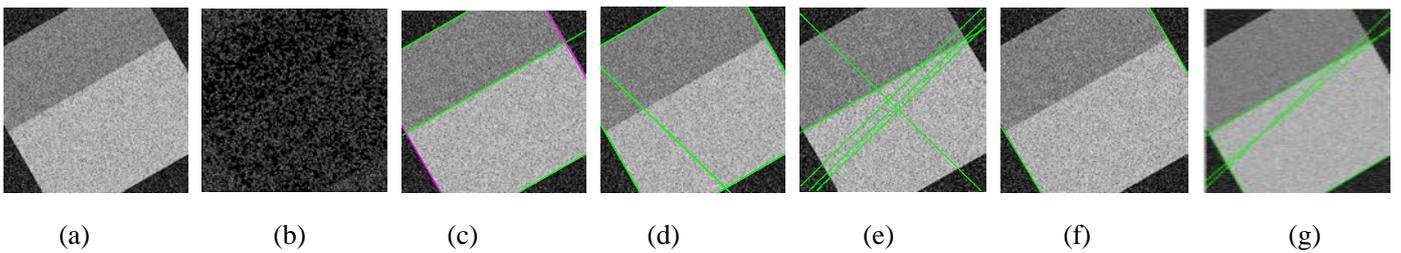

(a)　　　(b)　　　(c)　　　(d)　　　(e)　　　(f)　　　(g)

Fig. 11. Lines are detected from step image with additive salt and pepper noise. (a) Image suffers from salt and pepper noise with noise density 0.3. (b) Binary edge image obtained by Canny detector with parameters $[T_{low}, T_{high}, \sigma] = [0.14, 0.35, 1]$. Lines extracted by (c) funnel transform, (d) KHT, (e) SHT, (f) LSD and (g) FHT.

**1. Effect of additive noise on detection performance:** The performance of the straight line detection method is affected by many factors, in which the influence of noise should not be ignored and is unavoidable. Figs. 10 and 11 compare the robustness of our detection method, KHT, SHT, FHT and LSD on different types of noise. If several short lines are on the same long line, LSD usually reports each short lines rather than the long one. But in this experiment, if LSD detects two or more short lines from the same long straight line, only the long line is reported. For the sake of a clear sight of false detection, all methods involved in comparison, except LSD, do not adopt line verification step. The line verification step included in LSD is used to extract lines rather than to remove false detection, which ensures LSD to work normally. Furthermore, to be fair, we have removed the salient (peaks) points coming from the boundaries aroused by expanding the input image. Fig. 10 displays a synthetic image added with Gaussian white noise. The additive Gaussian white noise is zero mean noise with variance 0.1. Fig. 11 shows an image added with salt and pepper noise. The noise density of additive salt and pepper noise is 0.3. In Figs. 10 and 11, the images from left to right denote the noisy image and the different results of line detection using the funnel transform, KHT, SHT, LSD and FHT, where the denoted detection results have been overlaid on the noise image. In the detection results obtained by the funnel transform, all the green lines are detected from the regular space and all the carmine lines come from the inverse space. The detection results shown in Figs. 10 and 11 illustrate that the funnel transform can extract all the straight lines. KHT finds most correct lines but is affected by noise, and extracts one false line. SHT receives the largest number of false lines. LSD does not declare any false detection, but it misses some true lines. Under the Gaussian white noise case, FHT clearly declares all the true lines, while under the case of salt and pepper noise, the performance of FHT is decreased. LSD shows two false lines. The results given in Figs. 10 and 11 illustrate that the FT-SLD method is robust to noise.

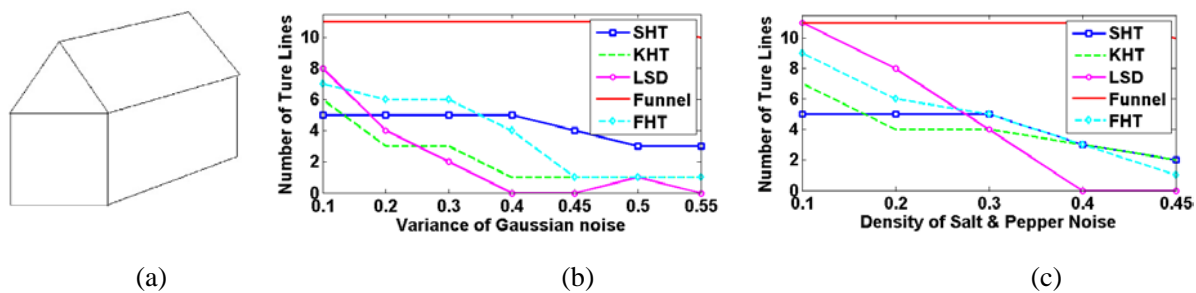

(a)　　　　　　　　　　　　　　(b)　　　　　　　　　　　　　　(c)

Fig. 12. The influence of noise intensity to different detectors is denoted. (a) An image with 11 lines. (b) Detection performances versus variance of Gaussian noise. (c) Detection performances versus density of salt and pepper noise.

To evaluate how far the proposed method goes on robustness to noise, we change the intensity of the added noise until the first false line arises in the detection result of funnel transform. Fig. 12 shows the curves of the number of true lines versus noise intensity. In this experiment, the benchmark image reported in [42] is used, as given Fig. 12(a). This image of a house consists of 11 straight lines. Fig. 12(b) compares the detection results obtained by the funnel transform and some other representative methods, KHT, SHT, LSD and FHT, where they process the image disturbed by zero mean Gaussian white noise. Fig. 12(c) compares the detection results by processing the image impaired by salt and pepper noise. Every method declares at most 11 most-relevant lines. In this experiment, different detection methods show different anti-noise capabilities. When the intensity of noise is small, some long edges are disconnected. These true disconnected edges do not largely affect the detection results of KHT, SHT, LSD and FHT. When noise variance is increased, the number of false edges is increased. The robustness of the funnel transform benefits from directly processing grayscale image. The experimental results demonstrate that the funnel transform has good robustness against noise, especially in Gaussian noise environments.

**2. Effect of multiplicative noise:** Specially, a good result is also obtained by applying the funnel transform to an image subject to multiplicative noise. Different from additive noise, the existence of multiplicative noise remarkably increases the complexity of the task for finding an edge pixel. Edge filters designed for optical images detect many false edge-pixels in the presence of the multiplicative noise, while edge filters designed for speckle images detect many distorted and biased edges. Accordingly, most of the commonly-used straight line detection methods will lose their effectiveness, because their detection abilities are heavily limited by the performance of an edge filter. Fig. 13 shows the detection result of the funnel transform on a speckle image with the same scene as in Fig. 12(a).

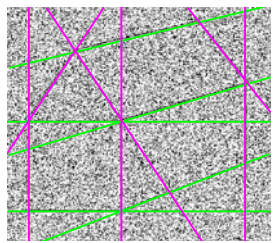

Fig. 13. Detected result of funnel transform on Fig. 12(a) with multiplicative noise, where green lines are obtained from the regular parameter space. Note that since Canny operator is not suitable to handle any image with multiplicative noise, the other methods based on the binary edge images do not get satisfactory line detection results.

**3. Effect of line thickness:** Except for the noise, lines with different thickness can be viewed as another challenge for the straight line detection method. The test man-made image shown in Fig. 14 includes several lines with thickness greater than one pixel. The detection results of the funnel transform, KHT, SHT, LSD and FHT for those lines in Fig. 14(a) are shown in Fig. 14(b), (c), (d), (e) and (f), respectively. As shown in Fig. 14(b), the funnel transform can accurately obtain the position of all the lines. For the line with more than one pixels width, the funnel transform reports two parallel straight lines at two sides of the line. From Fig. 14(c), (d) and (e), it can be easily seen that only the lines with one pixel are correctly extracted by KHT, SHT and LSD. FHT can correctly extract lines with single and three pixels. For the lines with five pixels width, KHT, SHT and FHT give several intersecting lines. If increasing the resolution of accumulator cell, they will correctly report two parallel lines, which also increase the computational cost. The detection results of LSD are quite different from that of SHT. LSD easily verifies the position of the lines having several pixels width, but it cannot correctly report the position of the lines having single pixel width, because for the line with single pixel width, the gradient information on the two sides of the line is different.

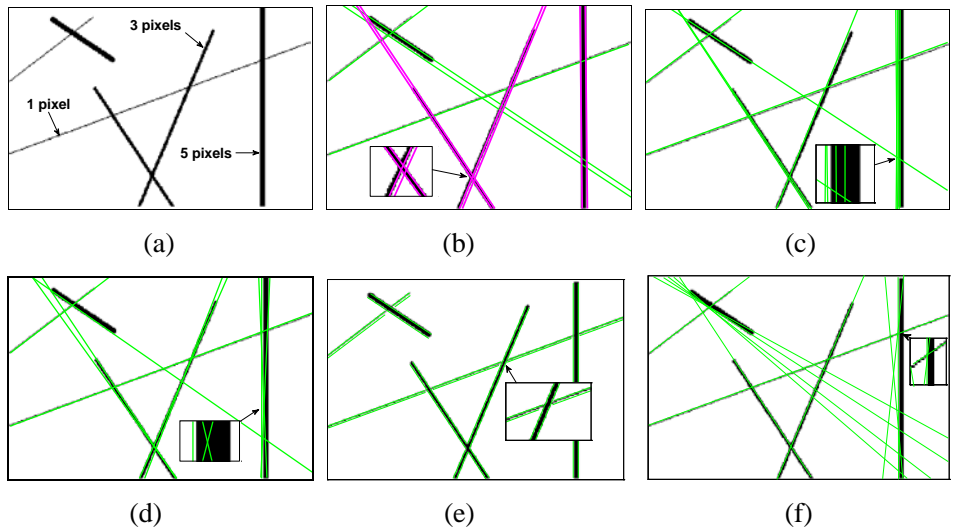

Fig. 14. Line detection methods are used to detect the lines with different thickness. (a) Image of lines with six different thickness. (b) Lines detected by funnel transform. (c) Lines detected by KHT. (d) Lines detected by SHT. (e) Lines detected by LSD. (f) Lines found by FHT.

**4. Effect of occlusion:** Mutual occlusion in the shapes is a common phenomenon because an actual image is complex. For example, intersection of lines is a special case of occlusion. Mutual occlusion could affect the results of some line detection methods. Fig. 15 shows an example of the image with 8 intersecting lines which are occluded by a disk. The image is a square with size $351 \times 351$. The diameter of occluded disk in the first, second, third, and fourth rows of Fig. 15 is 0, 77, 129 and 286 pixels, respectively. The funnel transform, KHT, SHT and FHT display the 8 most-relevant detected lines, while LSD shows all the detected short lines. As the diameter of the disk increases, different detection methods reveal different anti-occlusion abilities. The experiment results given in Fig. 15 reveal that the funnel transform can focus on all the 8 true lines for the images having the occluded disk with different diameters. When the occluded diameter is equal to 286 pixels, SHT begins to declare the first one false line. In such a situation, FHT shows three false lines. As the occluded diameter increases, the performance of KHT is going to be worse. That is because the presence of occlusion affects the cluster algorithm of KHT. When the area of occlusion is enlarged, the cluster grouping algorithm cannot correctly identify clusters of approximately collinear features pixels. Consequently, some lines will not be correctly detected by KHT if serious occlusion is present. LSD is also sensitive to occlusion. Even for two intersecting long lines, they will be interpreted as 4 short lines. As a consequence, LSD declares 16, 31, 37, and 46 lines for the occluded diameter of 0, 77, 129 and 286 pixels, respectively. For the bigger occluded diameter, the boundaries of the occluded disk are accepted by LSD as several short straight lines. Hence, the number of lines detected by LSD is much greater than 16. However, the boundaries of the disk are barely locally approximated as line structures since they are not real straight lines. It is unacceptable that an approximate line structure is interpreted as a straight line. We notice that KHT and LSD both adopt an algorithm of grouping edge-pixels. Although the grouping algorithms are different in KHT and LSD, the purpose of the grouping algorithms is to group pixels into several groups with similar orientation. Mutual occlusion disturbs grouping algorithms. Hence, the anti-occlusion abilities of KHT and LSD are not very good. Fig. 15 shows that the funnel transform and SHT have good anti-occlusion abilities, in which the funnel transform is better.

| Original image | Funnel transform | KHT | SHT | LSD | FHT |

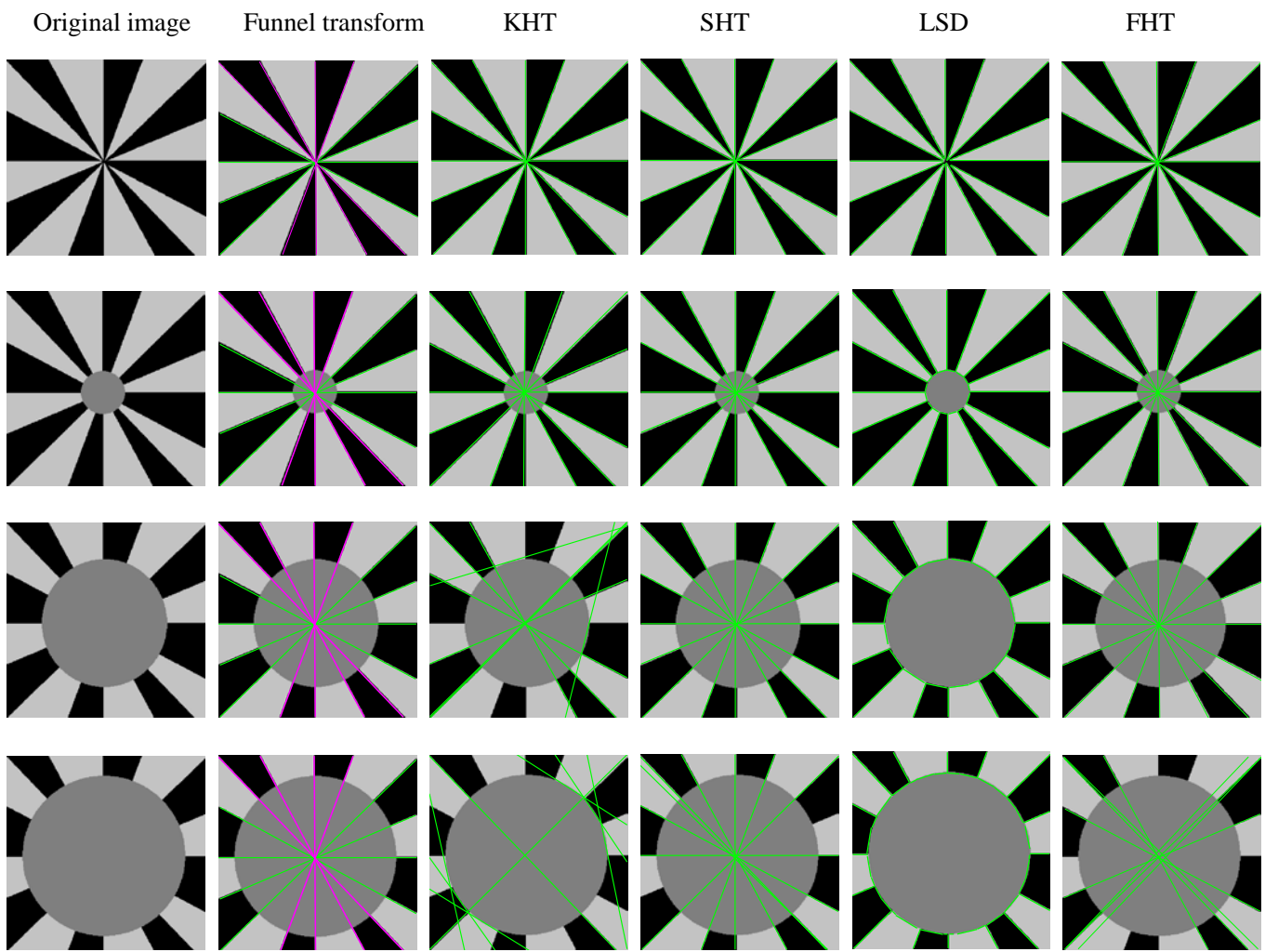

Fig. 15. Anti-occlusion ability of four methods is shown. Every image has size $351 \times 351$ and includes 8 lines. From the first to the fourth row, the diameters of occlude circles are 1, 77, 129 and 286 pixels, respectively. The first column is the original image. From the second to sixth column, the detection result of the funnel transform, KHT, SHT, LSD and FHT are denoted, respectively.

**5. Effect for processing long lines:** Fig. 16 shows a comparison of long line detection ability of the funnel transform, KHT, SHT and FHT on a series of natural images. LSD does not get into comparison since it is well adapted for short line detection but not for long line detection. The funnel transform, KHT, SHT and FHT are performed on the images of Water Cube [43], church [44], wall [21] and building [45]. All the four images have abundantly geometrical contents. Specifying the number of the detected lines is a usual application of the straight line detection method. According to the complexity of these natural images, we specify different numbers of the detected lines.

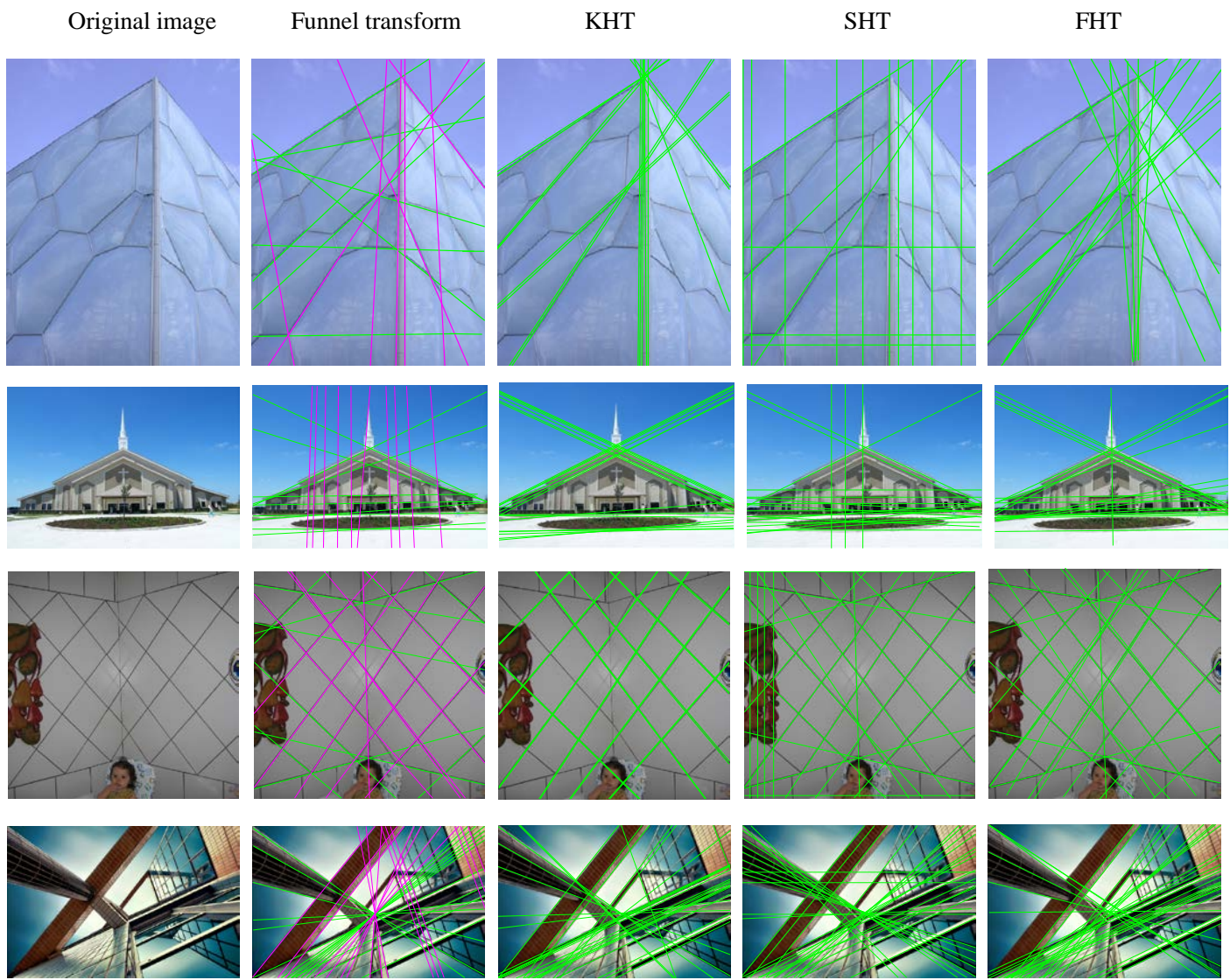

Fig. 16. Performance comparisons of the funnel transform, KHT, SHT and FHT on real images. The images are arranged in the scene complex degree. The simpler the scene, the smaller number of the most-relevant lines is extracted. We respectively display 16, 20, 22 and 30 most-relevant lines from Water Cube, Church, Wall and Building. From the first to the fifth column, these images correspond to the original image, the detection results of the funnel transform, KHT, SHT and FHT, respectively.

For most images, KHT can report the majority of important lines. However, when blurry edges or small gradients are present, KHT would produce many variants (alias) of the same lines. This situation is clearly evident in the church image where multiple concurrent lines corresponding to the roof are extracted by KHT. The shadow of the roof makes the gradients small, and hence leads KHT to report multiple parallel lines. Although KHT is able to identify most of the important lines in all the natural images, it often loses some of important lines. When the number of the detected lines is fixed, too many parallel lines are reported. For example, KHT does not implement the detection for the diagonal lines on the four corners of the wall image and the tile column boundary of the building image. KHT detects too many parallel lines since its grouping algorithm fails to group neighboring edge-pixels into appropriate clusters. The voting scheme of SHT is sensitive to the texture of the image. Some hallucinatory aligned texture structures make SHT bring about some inappropriate line reports. These lines caused by the texture are obviously meaningless and hence are viewed as false detection. For example, the false lines are caused by the dimly visible texture of Water Cube and the false lines caused by the grass in the church image. FHT can report most of important lines in all the natural images. However, as the complex of the scene increases, some of important lines will not be identified as shown in the beam of the building). This is because the important line structures may be drowned in the amounts of the textural of the image. The funnel transform has a good detection performance for most cases, but it is difficult to perfectly extract some short lines. At this point, KHT exhibits a better detection result (see the boundaries of the small beam on the top of the building image). The cluster grouping algorithm of KHT skillfully makes use of some pixel information of the image, which leads to KHT being sensitive to short lines. A limitation of the funnel transform is the lack of detection of some short lines, since the funnel transform mainly focuses on the global information of the image.

## VII. Conclusion

We have proposed a new transform method named as funnel transform which can be used to quickly extract straight lines directly from the optical and speckle images. The funnel transform can work without depending on an edge detector, and can directly apply to grayscale images, since its nonlinear variable-metric transform (NVMT) has the high-pass filtering ability. The funnel transform maps image lines into salient (peaks) points. Each salient point is theoretically described by a 2D delta function. Consequently, the position of a line in the image can be easily determined by the coordinates of the salient (peaks) point in the parameter space. The detection precision and the computational complexity of the funnel transform are only associated with the size of the input image. When the number of lines to be detected is specified, the funnel transform can accurately detect the lines in different images without manually tuning parameter for each image. In addition, the funnel transform generally involves lower computational complexity than the standard Hough transform and the fast implementation of Radon transform. The funnel transform not only works normally under the case with small noise, but also exhibits a good robustness against heavy noise. When the true lines are disconnected by mutual occlusion and intersection, the funnel transform is capable of identifying all the lines properly. More importantly, the funnel transform is less sensitive to the image texture. However, every method has its own advantages and disadvantages, and the funnel transform has no exception. The funnel transform is insensitive to true short lines. Hence, in real images the boundaries of some tiny structures will be missed out. Dividing the image into multiple smaller ones can improve the performance of short line detection, but dividing the image will increase the computational complexity and decrease the detection precision for the funnel transform as well. If we want to simultaneously detect long and short lines, then the global information of the input image should be used along with the local information. In our future work, we will study how to fully exploit image information and to effectively identify short lines. Besides, we are also trying to extend the funnel transform method in order to detect arbitrary curves and shapes.

## Appendix A: Theory of Funnel Transform in High dimensional Space

It is well-known that in the three-dimensional space, an ideal point image can be described by 3D delta function

$$I_g(x,y,z) = \delta(x-x_0)\delta(y-y_0)\delta(z-z_0). \tag{A.1}$$

It is directly tested that the ideal point image (A.1) equals infinity and zero at $(x_0, y_0, z_0)$ and in other places, respectively. Using the idea similar to the funnel transform on a 2D point, we first do 1D Fourier transform to the independent variable $z$ of $I_g(x,y,z)$ to obtain

$$\begin{aligned}\bar{I}(x,y,\omega_3) &= \int_{-\infty}^{+\infty} I_g(x,y,z)e^{-j\omega_3 z}dz = \int_{-\infty}^{+\infty} \delta(x-x_0)\delta(y-y_0)\delta(z-z_0)e^{-j\omega_3 z}dz \\ &= \delta(x-x_0)\delta(y-y_0)\int_{-\infty}^{+\infty}\delta(z-z_0)e^{-j\omega_3 z}dz \\ &= \delta(x-x_0)\delta(y-y_0)e^{-j\omega_3 z_0}\end{aligned} \tag{A.2}$$

Then, substituting $\omega_3 x = \omega_{max} x'$ and $\omega_3 y = \omega_{max} y'$ into equation (A.2) yields

$$\bar{I}(x',y',\omega_3) = \delta(x'-(\omega_3/\omega_{max})x_0)\delta(y'-(\omega_3/\omega_{max})y_0)e^{-j\omega_3 z_0}. \tag{A.3}$$

With respect to the independent variables $x'$ and $y'$, doing 2D Fourier transform of $\bar{I}(x',y',\omega_3)$ leads to

$$\begin{aligned}
\overline{I}(\omega_1,\omega_2,\omega_3) &= \int_{-\infty}^{+\infty}\int_{-\infty}^{+\infty} \overline{I}(x',y',\omega_3)e^{-j\omega_1 x'}e^{-j\omega_2 y'}dx'dy' \\
&\quad \int_{-\infty}^{+\infty}\int_{-\infty}^{+\infty} \delta(x'-(\omega_3/\omega_{max})x_0)\delta(y'-(\omega_3/\omega_{max})y_0)e^{-j\omega_3 z_0}e^{-j\omega_1 x'}e^{-j\omega_2 y'}dx'dy' \\
&= \exp(-j\omega_1 \frac{\omega_3}{\omega_{max}}x_0)\exp(-j\omega_2 \frac{\omega_3}{\omega_{max}}y_0)\exp(-j\omega_3 z_0) \\
&= \exp\left[-j(z_0+\frac{\omega_1}{\omega_{max}}x_0+\frac{\omega_2}{\omega_{max}}y_0)\omega_3\right]
\end{aligned} \quad (A.4)$$

Finally, doing 1D inverse Fourier transform for the independent variable $\omega_3$ in equation (A.4) yields

$$\begin{aligned}
\overline{I}(\omega_1,\omega_2,z) &= \int_{-\infty}^{+\infty} \overline{I}(\omega_1,\omega_2,\omega_3)e^{j\omega_3 z}d\omega_3 \\
&= \int_{-\infty}^{+\infty} \exp\left[-j(z_0+\frac{\omega_1}{\omega_{max}}x_0+\frac{\omega_2}{\omega_{max}}y_0)\omega_3\right]e^{j\omega_3 z}d\omega_3 \\
&= \delta(z-\frac{x_0}{\omega_{max}}\omega_1-\frac{y_0}{\omega_{max}}\omega_2-z_0)
\end{aligned} \quad (A.5)$$

We can verify that the 1D delta function (A.5) in 3D space indicates an ideal plane image that is equal to infinity and zero in and without the plane $z-(x_0/\omega_{max})\omega_1-(y_0/\omega_{max})\omega_2-z_0=0$, respectively. Very interestingly, funnel transform let a 3D ideal point image have been mapped into a 3D ideal plane image.

Conversely, an ideal plane image in 3D space can be represented by a 1D delta function

$$s(x,y,z) = \delta(z-ax-by-c). \quad (A.6)$$

Note that $s(x,y,z)$ equals infinity and zero in and without the 3D plane $z-ax-by-c=0$, respectively. Adopting the idea similar to the funnel transform, we first do 1D Fourier transform of $s(x,y,z)$ with respect to the independent variable $z$, and obtain

$$\begin{aligned}
F(x,y,\omega_2) &= \int_{-\infty}^{+\infty} s(x,y,z)e^{-j\omega_3 z}dz = \int_{-\infty}^{+\infty} \delta(z-ax-by-c)e^{-j\omega_3 z}dz \\
&= \exp(-j\omega_3(ax+by+c)) = \exp(-ja\omega_3 x)\exp(-jb\omega_3 y)\exp(-j\omega_3 c))
\end{aligned} \quad (A.7)$$

Then, substituting $\omega_3 x = \omega_{max} x'$ and $\omega_3 y = \omega_{max} y'$ into equation (A.7) yields

$$F(x',y',\omega_3) = e^{-ja\omega_{max}x'}e^{-jb\omega_{max}y'}e^{-j\omega_3 c}. \quad (A.8)$$

With respect to three independent variables $x'$, $y'$ and $\omega_3$, performing 3D inverse Fourier transform of $F(x',y',\omega_3)$ gives within a constant

$$F(\omega_1, \omega_2, z) = \int_{-\infty}^{+\infty}\int_{-\infty}^{+\infty}\int_{-\infty}^{+\infty} F(x', y', \omega_3) e^{j\omega_1 x'} e^{j\omega_2 y'} e^{j\omega_3 z} dx' dy' d\omega_3$$

$$= \int_{-\infty}^{+\infty}\int_{-\infty}^{+\infty}\int_{-\infty}^{+\infty} e^{-ja\omega_{max} x'} e^{-jb\omega_{max} y'} e^{-j\omega_3 c} e^{j\omega_1 x'} e^{j\omega_2 y'} e^{j\omega_3 z} dx' dy' d\omega_3 \quad . \tag{A.9}$$

$$= \int_{-\infty}^{+\infty} e^{j(\omega_1 - a\omega_{max})x'} dx' \int_{-\infty}^{+\infty} e^{j(\omega_2 - b\omega_{max})y'} dy' \int_{-\infty}^{+\infty} e^{j\omega_3(z-c)} d\omega_3$$

$$= \delta(\omega_1 - a\omega_{max})\delta(\omega_2 - b\omega_{max})\delta(z - c)$$

So far an ideal plane image has been mapped into an ideal point (peak) image in 3D space.

***Remark A.1:*** *Comparing (A.5) and (A.9), we know that the funnel transform can achieve the bidirectional mapping between the 3D plane and the point in 3D space. This is because both an ideal plane image and an ideal plane are determined by three parameters (see expressions (A.1) and (A.6)).*

***Remark A.2:*** *In 3D space, an ideal line image is represented by a 2D delta function*

$$I(x, y, z) = \delta(a_1 x + a_2 y + a_3 z + a_4)\delta(b_1 x + b_2 y + b_3 z + b_4) . \tag{A.10}$$

*It is easily verified that this ideal line image is respectively equal to infinity and zero on and without the line* $\begin{cases} a_1 x + a_2 y + a_3 z + a_4 = 0 \\ b_1 x + b_2 y + b_3 z + b_4 = 0 \end{cases}$ *. It is seen from (A.10) that an ideal line image seems to have six parameters. In fact, a line includes at most four parameters (two intercepts and two angles) in which two angle parameters determine the direction of the line. Unfortunately, in 3D space, there is generally not any mapping between an ideal line image and an ideal point image, because the parameters of the ideal line image are over those of the ideal point image. Fortunately, in 3D space, every two peaks in 3D parameter space can determine a line in 3D space, since each peak corresponds a plane image and two nonparallel planes ensure a line. Thus, in 3D space, we can indirectly detect the line images by detecting the 3D peaks in 3D parameter space.*

***Remark A.3:*** *More generally speaking, it is not difficult that the funnel transform in 3D space is extended to case in N-dimensional space ($N \geq 4$). We believe that there is a bidirectional mapping between a plane and a point in N-dimensional space.*

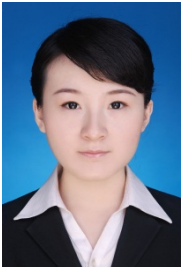
Qian-Ru Wei was born in Shaanxi province, China, on October 18, 1986. She received the B.S. degree from the department of Network Engineering, Nanjing University of Science and Technology, Nanjing, China, in 2009. She is currently pursuing the Ph.D. degree at Xidian University, Xi'an, China.

Her main research interests are in edge detection and corner feature extraction of SAR image.

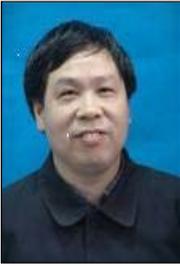
Da-Zheng Feng was born in December 1959. He received the Diploma degree from Xi'an University of Technology, Xi'an, China, in 1982, the M. S. degree from Xi'an Jiaotong University, Xi'an, China, in 1986, and the Ph.D. degree in electronic engineering from Xidian University, Xi'an, China, in 1995. From May 1996 to May 1998, he was a Postdoctoral Research Affiliate and an Associate Professor with Xi'an Jiaotong University, China. From May 1998 to June 2000, he was an Associate Professor with Xidian University. Since July 2000, he has been a Professor at Xidian University. He has published more than 200 journal papers. His current research interests include signal processing, intelligence and brain information processing, image processing, artificial neural networks, and pattern recognition. He is a member of IEEE.

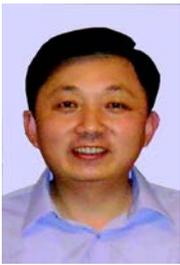
Wei-Xing Zheng received the Ph.D. degrees in Electrical Engineering from Southeast University, China in 1989. He has held various faculty/research/visiting positions at Southeast University, China, Imperial College of Science, Technology and Medicine, UK, University of Western Australia, Curtin University of Technology, Australia, Munich University of Technology, Germany, University of Virginia, USA, and University of California-Davis, USA. Currently he holds the rank of Full Professor at University of Western Sydney, Australia. Dr. Zheng has served as an Associate Editor for a number of flagship journals, including IEEE Transactions on Circuits and Systems-I: Fundamental Theory and Applications (2002--2004), IEEE Signal Processing Letters (2007--2010), IEEE Transactions on Circuits and Systems-II: Express Briefs (2008--2009), IEEE Transactions on Automatic Control (2004--2007 and 2013--now), Automatica (2011--now), IET Control Theory and Applications (2013--now), and IEEE Transactions on Fuzzy Systems (2014--now). He was a Guest Editor of Special Issue on Blind Signal Processing and Its Applications for IEEE Transactions on Circuits and Systems-I: Regular Papers (2009--2010). He has also served as the Chair of IEEE Circuits and Systems Society's Technical Committee on Neural Systems and Applications and as the Chair of IEEE Circuits and Systems Society's Technical Committee on Blind Signal Processing. He is a Fellow of IEEE.